\documentclass{Interspeech}

\usepackage{cite}
\usepackage{amsmath,amssymb,amsfonts}
\usepackage{algorithmic}
\usepackage{graphicx}
\usepackage{subfigure}
\usepackage{multirow}
\usepackage{textcomp}
\usepackage{xcolor}
\usepackage{hyperref}
\usepackage{tikz}
\usepackage{pgfplots}
\usepackage{booktabs}
\usepackage{csvsimple}
\usepackage{array}
\usepackage{subcaption}
\usepackage{soul}

\definecolor{mygray}{gray}{0.9}
\interspeechcameraready

\title{Enhanced Hybrid Transducer and Attention Encoder Decoder with Text Data}

\author[]{Yun}{Tang}
\author[]{Eesung}{Kim}
\author[]{Vijendra}{Raj Apsingekar}

\affiliation[nocounter]{}{Samsung Research America}{USA}
\email{y.tang@samsung.com}
\keywords{Multimodality learning, Multi-task learning, Speech recognition, Domain adaptation}

\usepackage{comment}
\begin{document}

\maketitle

\begin{abstract}    
%A joint speech and text optimization method is proposed for hybrid transducer and attention-based encoder decoder (TAED) modeling to leverage large amounts of text corpus and enhance automatic speech recognition (ASR) accuracy. 
%The joint TAED (J-TAED) is trained with both speech and text input modalities together, while it only takes speech data as input during inference. 
%J-TAED is optimized under a multitask learning framework with two main subtasks, i.e., the attention-based encoder decoder task and the Transducer task. 
%The trained model can unify the internal representations from different modalities, and can be further extended to text based domain adaptation.  It can effectively alleviate data scarcity issues during domain adaptation since no speech data is required.  
%Our experimental results show J-TAED could integrate speech 
%and linguistic information into one model. Compared with the strong baseline trained with speech data only, the jointly trained system can reduce the WER by 5.8 $\sim$ 12.8\% relative to the \textsc{Librispeech} dataset. The model is also evaluated on two out-of-domain datasets, one is financial data set \textsc{SPGISpeech} and another is a named entity focused In-House dataset. The text based 
%domain adaptation brings 15.3\%  and 17.8\% relative WER reduction on those two datasets respectively. 
A joint speech and text optimization method is proposed for hybrid transducer and attention-based encoder decoder (TAED) modeling to leverage large amounts of text corpus and enhance ASR accuracy.
The joint TAED (J-TAED) is trained with both speech and text input modalities together, while it only takes speech data as input during inference. The trained model can unify the internal representations from different modalities, and be further extended to text-based domain adaptation.  It can effectively alleviate data scarcity for mismatch domain tasks since no speech data is required.  
Our experiments show J-TAED successfully integrates speech and linguistic information into one model, and reduce the WER by 5.8 $\sim$ 12.8\% on the \textsc{Librispeech} dataset. The model is also evaluated on two out-of-domain datasets: one is finance and another is named entity focused. The text-based domain adaptation brings 15.3\%  and 17.8\% WER reduction on those two datasets respectively.
\end{abstract}

\section{Introduction}

The neural based end-to-end framework becomes the mainstreaming in automatic speech recognition (ASR)~\cite{Li2021RecentAI}. Many works are devoted to enhance the modeling ability~\cite{Graves2006ConnectionistTC,Graves2012SequenceTW,Gulati2020ConformerCT,Tang2023HybridTA} and improve the data efficiency~\cite{park2019specaugment,Bapna2021SLAMAU,Tang2022UnifiedSP,Zhang2022SpeechUTBS}. 

The end-to-end modeling frameworks can roughly divided into two categories, label synchronized methods, such as Attention based Encoder-Decoder modeling (AED)~\cite{Bahdanau2014NeuralMT}, and time synchronized methods, such as connectionist temporal classification (CTC)~\cite{Graves2006ConnectionistTC} and Transducer~\cite{Graves2012SequenceTW}. Compared with the label synchronized decoding method, time synchronized decoding methods are less impacted by the hallucination effect since every output token generated is conditioned on one specific input frame/token.  It also makes methods in this category naturally compatible with streaming applications. 
On the other hand, label synchronized decoding, such as AED, doesn't have limitation to access encoder output 
information, and it can make predictions based on the entire input sequence instead of one of them. Though it might suffer from over-generation or under-generation issues during decoding. Hybrid transducer and attention-based encoder decoder (TAED)~\cite{Tang2023HybridTA} attempts to integrate both methods into one model and shows encouraging 
improvement.

Compared with the conventional ASR system~\cite{Young1995TheHB}, there is no dedicated acoustic model (AM) and language model (LM) in end-to-end ASR models. The end-to-end model usually is trained with speech training dataset directly and text-based training corpus is not used, which is used to build dedicated LM to integrate linguistic information in the conventional ASR system.
It poses a great challenge to incorporate large amounts of text data for end-to-end ASR models. 
Many efforts have done to leverage large amount of text training data
into ASR model build~\cite{Tang2020AGM,Bapna2021SLAMAU,Chung2021w2vBERTCC,Tang2022UnifiedSP,Zhang2022SpeechUTBS}. Work \cite{Bapna2021SLAMAU,Chung2021w2vBERTCC,Chen2022MAESTROMS} attempt to integrate linguistic information into 
encoder, while \cite{Tang2020AGM, Tang2022UnifiedSP, Zhang2022SpeechUTBS} focus on attention-based encoder and decoder model and encode linguistic information into decoder. 

In this work, we focus on leveraging text data to enhance TAED performance. We propose J-TAED, which jointly optimizes text and speech input modalities during training.  
A multimodality encoder is proposed to alleviate the modality mismatch.
The text input modality could be learned with paired speech input, and it can be trained alone.  
It makes J-TAED being able to leverage large amounts of text corpus during training and enhance the
recognition accuracy.
The J-TAED model generates unified representations for speech and text modalities, and is further 
extended to text-based adaptation for out-of-domain tasks. 
Our results show J-TAED can 
effectively utilize the text data to enhance the ASR performance and alleviate domain mismatch.
The main contributions of this work include:
\begin{enumerate}
    \item Propose J-TAED, a multimodality based TAED model for ASR
    \item Propose a multimodality encoder to alleviate the modality mismatch
    \item Propose text-based domain adaptation to improve the performance on the out-of-domain tasks when the speech training data is not available  
\end{enumerate}
%The paper is organized as below.  \autoref{sec:methods} describes the proposed joint speech text training framework, \autoref{sec:related_work} discusses the related work, the experiments are presented in~\autoref{sec:expt} and the last section draws the conclusion.

\begin{figure}[t!]
\centerline{\includegraphics[width=0.7\columnwidth]{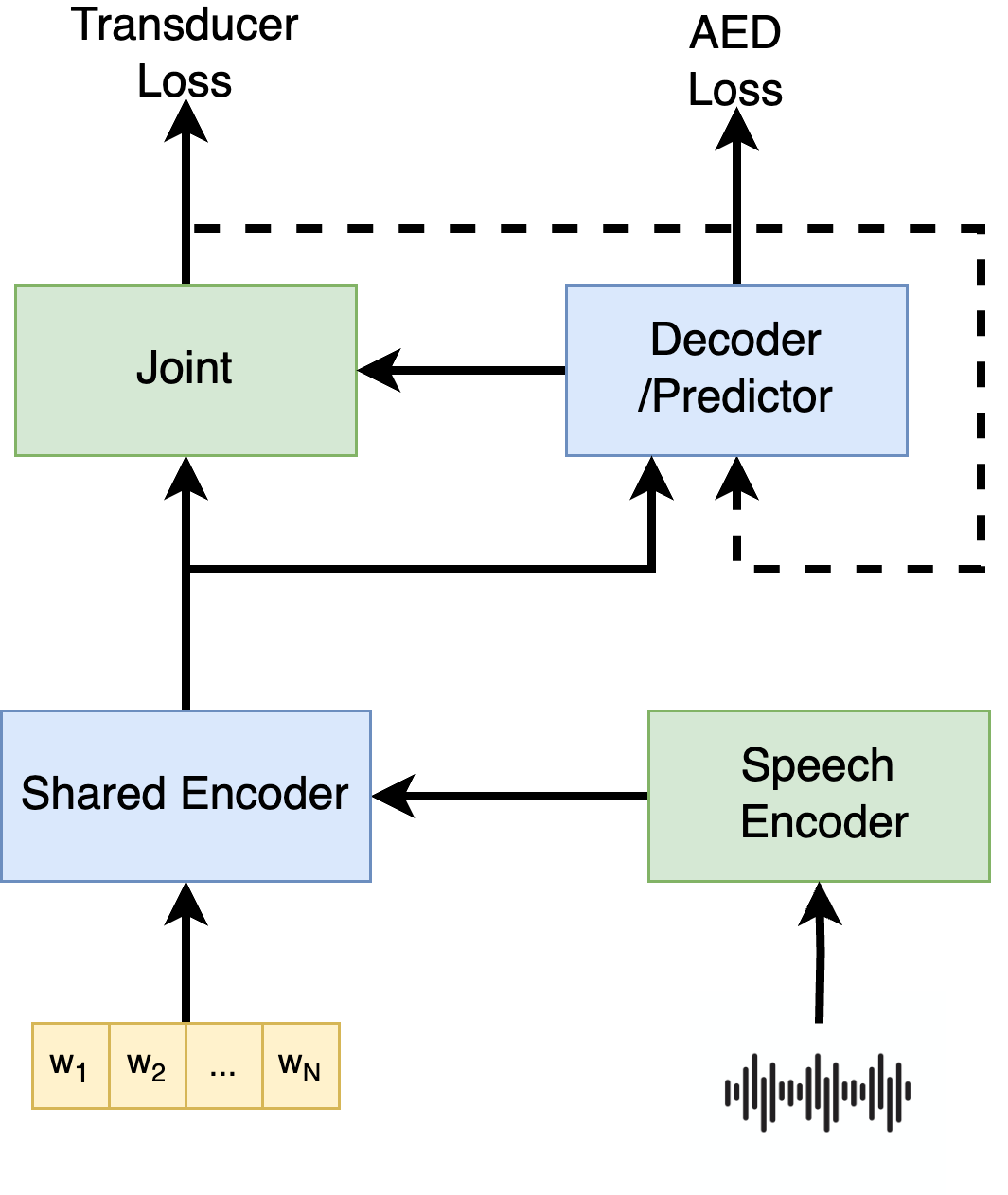}}
\caption{Joint speech text optimization for TAED. The blue blocks are shared by both speech and text modalities.}
\label{fig:j-taed}
\end{figure}

\section{Methods}\label{sec:methods}

\subsection{Joint Speech Text TAED}\label{sec:j-taed}
TAED~\cite{Tang2023HybridTA} is a hybrid framework to integrate both Transducer and AED into one model for speech recognition and translation. The model is optimized with Transducer criterion~\cite{Graves2012SequenceTW} and cross entropy criterion for AED. Assume $D_s=\{(x_s,y)\}$ is a speech data set, where $x_s$ is speech input and $y$ is corresponding transcript, the optimization loss is 
\begin{equation}
    \mathcal{L}_{TAED} = \sum_ {(x_s, y) \in D_s}\mathcal{L}_{TR}(x_s|y) + \mathcal{L}_{AED}(x_s|y).
\end{equation}
where $\mathcal{L}_{TR}(x_s|y)$ and $\mathcal{L}_{AED}(x_s|y)$ are Transducer loss from joint outputs and cross entropy loss from decoder (predictor) outputs.

The proposed J-TAED method extends TAED with multimodality inputs, as shown in \autoref{fig:j-taed}. The new approach consumes both speech and text inputs during training 
and takes speech input only during inference.  The text inputs come from two sources. First is the transcript of the corresponding speech data, and second is from the unpaired text data $D_t(y)$. Following~\cite{Tang2020AGM}, we convert the text tokens into phoneme tokens to make the text input close to the speech input and mask part of the phoneme tokens to force the model to learn corresponding context information from the text-based input. The final text input to the model is $M(f_p(y))$, where $M$ is masking operation on the phoneme and $f_p$ is the conversion from text tokens to the corresponding phoneme tokens.

The model with text input is optimized with cross entropy on the decoder outputs  through the AED sub-task. We don't apply the Transducer criterion with text input due two reasons. First the resolution of the encoder outputs for the speech and text inputs are different, which leads to inferior performance in our experiments. 
Second, we argue the majority of linguistic information is encoded in the encoder and decoder modules, while joint is less effective to store linguistic information due to simple structure and limited parameters.

In order to encourage the knowledge transfer from text to speech, online-knowledge distillation~\cite{Tang2021ImprovingST} is applied at the decoder outputs for the paired speech and text inputs. The optimization function for J-TAED is modified as below:
%\begin{align}
%    \mathcal{L} &= \sum_ {(x_s, y) \in D_s, y'\in D_t}\mathcal{L}_{TR}(x_s|y) + \alpha_s \mathcal{L}_{AED}(x_s|y) \nonumber\\
%    & + \beta_t\mathcal{L}_{TR}(M(f_p(y))|y) + \alpha_t\mathcal{L}_{AED}(M(f_p(y))|y) \nonumber\\
%    & + \beta_t\mathcal{L}_{TR}(M(f_p(y'))|y') + \alpha_t\mathcal{L}_{AED}(M(f_p(y')|y') \nonumber\\
%    & + \gamma D_{KL}(p(x_s|y)|p(f_p(y)|y))
%\end{align}

\begin{align}
    \mathcal{L} &= \sum_ {(x_s, y) \in D_s, y'\in D_t}\mathcal{L}_{TR}(x_s|y) + \alpha_s \mathcal{L}_{AED}(x_s|y) \nonumber\\
    & + \alpha_t\big(\mathcal{L}_{AED}(M(f_p(y))|y)+ \mathcal{L}_{AED}(M(f_p(y'))|y')\big) \nonumber\\
    & + \beta D_{KL}\big(p_{AED}(x_s|y)|p_{AED}(f_p(y)|y)\big)
\end{align}
where $\alpha_s$, $\alpha_t$  and $\beta$ are weights for different auxiliary tasks respectively. $D_{KL}$ is KL divergence applied to the decoder outputs from paired speech and text inputs. $p_{AED}$ is the prediction probability from the AED decoder outputs. No masking is applied to the input phoneme tokens when the decoder outputs are used for online-knowledge distillation.

\subsection{Multimodality Encoder}\label{sec:multi-encoder}
The encoder is separated into two sub-encoders, i.e., speech encoder and shared encoder as shown in~\autoref{fig:j-taed}. The speech encoder is based on the Conformer encoder as~\cite{Gulati2020ConformerCT} and is dedicated to the speech input. It is with a downsampling module to reduce the speech input frames by 4 times and followed by a module with stacked Conformer layers. The shared encoder is assembled with Transformer~\cite{Vaswani2017AttentionIA} layers. 

The parameters in the shared encoder and predictor are shared by both speech and text modalities. The main motivation to switch Conformer layers to the Transformer layers in the shared encoder is to reduce the modality discrepancy. Compared with the Transformer layer, the Conformer layer has an extra convolution module, which is designed to model local information. However, the speech input resolution is different from the one in the corresponding text representations, i.e., the average speech input sequence length (after downsampling) to the shared encoder is 2 to 4 times %facts on the train_other_500: ave ratio is 2.7, min_ratio is 1.4, and max_ratio is 31
longer than the corresponding phoneme sequence length. The different resolutions could interfere with the modality information fusion within convolution modules. On the other hand, the self-attention module in the Transformer or Conformer layer exchanges information among input tokens via similarity, and it is less sensitive for different modalities and resolutions as shown in~\cite{Tang2021ImprovingST}. 
In our experiments, we find an encoder consisting of Conformer layers and then followed by Transformer layers could achieve similar performance as a Conformer encoder with less parameters.

Self-attention with relative position embedding~\cite{Shaw2018SelfAttentionWR} is used in both Conformer and Transformer layers. We have dedicated relative position embeddings for speech and text modalities, but parameters in self-attention modules are shared among two modalities.

%\subsection{Fusion Decoding with AED and Transducer}\label{sec:fusion}
%In TAED, two tasks are optimized that we can choose either AED decoding or Transducer decoding for inference. As mentioned in~\autoref{sec:intr}, Transducer adopts time synchronized decoding and is less impacted by the hallucination issue 
%while AED has the advantage to leverage audio information since it can access all encoder outputs.    
%Inspired by~\cite{Guo2024EffectiveIL}, we explore to fuse two decodings to enhance the decoding accuracy as below,
%\begin{align}
%    \log  p(t,y_i|x_s, y_{1:i-1}) &= (1-\mu) \log p_{TR}(t,y_i|x_s, y_{1:i-1})  \nonumber\\
%    & + \mu \log p_{AED}(y_i|x_s, y_{1:i-1})
%\end{align}
%where $y_i$ is a non-blank token, $p_{TR}$ is prediction probabilities from Transducer decoding outputs. Probabilities of blank tokens from Transducer are not changed.

\subsection{ASR Adaptation with text data}\label{sec:adapt}
%Adaptation is essential to adjust an ASR model trained on a general purpose dataset to out-of-domain tasks. 
It is essential to conduct adaptation to reduce mismatch when an ASR model is applied to an out-of-domain task. 
The mismatch mainly comes from two sources, acoustic variation and linguistic variation. The acoustic variation is due to the changes in environment, noise level, and speaking style etc, while the linguistic 
variation is highly related to the choices of words, phrases and topics.
The target domain text contains rich linguistic information and could be helpful to reduce linguistic mismatch. 
However, speech is the default input  and text is used as paired reference during training in conventional ASR models. 
It is challenge to utilize the target domain text directly in the conventional end-to-end ASR modeling. 

J-TAED learns speech and text modalities with shared parameters and unified representations.    
It makes the text-based adaptation possible. 
We explore to conduct text-based domain adaptation by optimizing 
the model with the AED loss %in TAED modeling 
using target domain text data, i.e., $\mathcal{L}_{AED}(M(f_p(y'))|y'))$, where $y'\in D_t$.
We hypothesis that the encoder in TAED is responsible for acoustic modeling while the predictor (decoder) is mainly in charge of linguistic modeling.
Hence, we only update parameters in the predictor in the text-based adaptation 
and keep parameters in other modules , i.e., encoder and joint, intact. %then the adapted model is used for decoding the target domain speech in the evaluation. 
We verify the text-based adaptation on two out-of-domain tasks and more details are shown in~\autoref{sec:expt}.

%The second approach, ``Partial-Decoder'', is to freeze parameters in the main decoder base in addition to the modules frozen in the first approach. The main decoder base includes the input embedding and Transformer layers, hence only layers after the last decoder Transformer layer could be updated. The advantage is that there is no parameter change for the TAED decoding, which could be used for the general purpose tasks, while the target domain application can be enhanced via AED decoding or fusion decoding aforementioned. 
%In this study, an extra linear layer with Layer Normalization is inserted between the last decoder Transformer layer and output embedding in the decoder, as shown in~\autoref{fig:j-taed}, to boost the performance of the ``Partial-Decoder''.  

\section{Related work}~\label{sec:related_work}
\noindent\textbf{Text-based adaptation} It is very attractive to adapt the trained End-to-End ASR models to out-of-domain datasets with text data only, however it is also challenging. Factorized neural Transducer (FNT)~\cite{Zhao2022FastAA,Wu2023TSOTFS,Liu2023ImprovedFN,Guo2024EffectiveIL} tackles this issue by predicting blank token and non-blank tokens with different predictors, hence a domain specific language model could be used to replace the non-blank predictor during inference. The adaptation results are promising, though minor accuracy degradation is observed in FNT compared with the vanilla Transducer model if no adaptation is applied. %In~\cite{Thomas2022IntegratingTI}, text tokens are converted to textogram features to match the audio durations. 
Our method is different from the previous methods that J-TAED is based on the TAED modeling. TAED outperforms the corresponding Transformer and J-TAED is more accurate than TAED. No compromise is required if we want to incorporate linguistic information into the ASR modeling. % and it outperforms both vanilla Transducer and TAED instead of an inferior version. Text based adaptation is further enhance J-TAED on the out-of-domain performance.
  
\noindent\textbf{Relation of Transducer and AED }Combining Transducer and AED might integrate advantages from both methods. One approach is through a two-pass approach~\cite{Sainath2019TwoPassES}.
An alternative is to replace the two-pass decoding with single-pass decoding, which integrates scores from CTC/Transducer with AED during the beam search~\cite{Watanabe2017HybridCA,Yan2022CTCAI,Sudo20224DAJ}.
Those approaches require dedicated decoders and there is no connection among those decoders. On the other hand, the AED decoder is also the predictor in TAED and any improvement on the AED decoder will transfer to TAED naturally, which 
lays the foundation for this work. 

\noindent\textbf{Integrating linguistic information in encoder} As discussed in~\autoref{sec:multi-encoder}, the paired speech and text feature sequences have different lengths and it poses a potential risk when model them with shared parameters. 
An up-sampling operation~\cite{Chen2022MAESTROMS,Thomas2022IntegratingTI,Huang2023TextonlyDA} is required  to process input phoneme sequences to align with corresponding speech input sequences. In this work, we proposed multimodality encoder and no up-sampling is required, which makes the joint training more efficient. 

\noindent\textbf{Speech large language model (LLM)} Decoder based speech LLM models~\cite{Chen2023SALMSL,Fathullah2023PromptingLL} are also beneficial from linguistic information encoded in LLM.  Compared with speech LLM, Transducer or TAED could support streaming applications naturally, and they are also able to achieve competitive results with less parameters since no LARGE language models are required~\cite{Chen2023SALMSL}.

\section{Experiments}\label{sec:expt}
\subsection{Datasets}
We conduct ASR modeling experiments on the 960 hours \textsc{Librispeech}~\cite{Panayotov2015LibrispeechAA} and the co-training text data is the language model training data\footnote{https://www.openslr.org/resources/11/librispeech-lm-norm.txt.gz} coming with the \textsc{Librispeech} dataset. 

% The text based domain adaptation experiments are conducted on the \textsc{SPGISpeech}~\cite{ONeill2021SPGISpeech50}, which includes 5000 hours transcribed finance teleconference audio data.
% We use the transcripts from the training set for the adaptation experiments and evaluate on the \textsc{SPGISpeech} testset.  There are about 1,926,805 utterances (46,301,311 words) in the training transcripts and 39,341 utterances in the testset. 

The text-based domain adaptation experiments are conducted on two datasets: the \textsc{SPGISpeech} 
dataset~\cite{ONeill2021SPGISpeech50}, which consists of financial audio data based on company earnings calls, and an In-House dataset dominated with English Named Entity (NE). 
For adaptation experiments, we utilize transcripts from the training set and evaluate performance on the
corresponding test sets. 
\textsc{SPGISpeech} comprises 5,000 hours of transcribed teleconference audio data. The training transcripts consist of approximately 1,926,805 utterances, encompassing 46,301,311 words, while the test set contains 39,341 utterances. 
The In-House data set includes a total of 824,868 utterances for adaptation. The dev and test sets each has 5 hours of speech data (5,236 utterances).
% we incorporate the Samsung In-House Named Entity (NE) Dataset to expand our text adaptation experiments. This dataset consists of English NE text data, with a total of 824,868 utterances. For evaluation purposes, we randomly select 5 hours of data (5,235 utterances) as the evaluation set and another 5 hours (5,236 utterances) as the test set, which is not used for training text data.

Target subword units are learned from SentencePiece~\cite{Kudo2018SentencePieceAS} 
 with vocabulary size 1024 and full character coverage on all \textsc{Librispeech} training text data. The grapheme to phoneme conversion for the input text is done through the ``g2p\_en'' Python package~\cite{Lee2018LearningPF}. The phoneme vocabulary size is 134.

\subsection{Experimental settings}

Input speech is represented as 80-dimensional log mel-filterbank coefficients computed every 10ms with a 25ms window. Utterance channel mean and variance normalization is applied. The SpecAugment~\cite{park2019specaugment} data augmentation is applied  with 2 frequency maskings of 27, and 10 time maskings at a maximum ratio of 0.05 in all experiments. 
The phoneme inputs are masked as whole word with ratio of 0.3.

%We choose the Transformer-Transducer (T-T)~\citep{Zhang2020TransformerTA} as our  model structure. 
As described in~\autoref{sec:multi-encoder}, there are two sub-encoders in our TAED/J-TAED.
The speech encoder starts with two 
convolution layers with a kernel size of three and a stride size of two. The input speech features are down-sampled by four and then processed by 11 
Conformer layers. The shared encoder has 6 Transformer layers. Both Conformer and Transformer layers are equipped with relative positional embedding~\cite{Shaw2018SelfAttentionWR}. Inter-CTC~\cite{Lee2021IntermediateLR} is applied on the 5th, 9th and 13th layers with weight 0.03 to boost the performance. For the joint training,  sub-task weights $\alpha_s$, $\alpha_t$  and $\beta$ in~\autoref{sec:j-taed} are set to 0.5, 0.3 and 0.6 respectively. 
%For the fusion decoding described in~\autoref{sec:fusion}, we set $\mu$ to 0.5, i.e., both Transducer and AED contributing equally in the final probability score.
There are 4 Transformer layers in the decoder (predictor) module.  
The Conformer/Transformer layers in both encoder and predictor have an input embedding size of 512, 8 
attention heads, and middle layer dimension 2048. The joiner module is a feed-forward neural network as 
Transformer Transducer~\cite{Zhang2020TransformerTA} except we insert a layer normalization~\cite{Ba2016LayerN} between two linear layers.  

The models are trained up to 300k updates using 8 A100 GPUs with NeMo~\cite{Kuchaiev2019NeMoAT}. 
The batch size is 240 seconds of audio and 10,000 tokens per GPU. 
Noam learning scheduler was used with a warmup of 10k steps and learning rate of 1.0. The encoder of TAED is initialized from trained Transducer model. 

In the adaptation experiments, the model is initialized from the J-TAED model built with \textsc{Librispeech} data and trained for another 20k steps %/home/jovyan/asr-sra-data/speechplus/work/eesung/librispeech/v3_cache/log/log_joint_text_taed_v2_dev_sp_1000_dbatch_lr.log (global step 18270)
with \textsc{SPGISpeech} text training data. The effective batch size is 160,000 tokens per GPU. %/home/jovyan/asr-sra-data/speechplus/work/eesung/librispeech/v3_cache/outputs/joint_text_taed.decoder_v2/share_dio.spgispeech_v2.lr0.01.se.aed0.0.taw1.wm.kl0.8.tr0.0.mr0.3.all_features.interctc0.05.noabsemb/2024-09-09_04-08-38/cmd-args.log

The Transducer optimization is based on Fast-RNNT~\cite{Kuang2022PrunedRF} for efficiency.
The best 5 checkpoints are averaged for the inference and we choose blank penalty~\cite{Tang2023HybridTA} $\tau=0.5$ for the TAED model. Beam size 4 is used in evaluation and the results are measured with word error rate (WER).

\subsection{Joint training}
The recognition results for the J-TAED are listed in~\autoref{tab:librispeech_rst}. The first two rows are the corresponding baselines for the Transducer and vanilla TAED trained from speech dataset only. The J-TAED results are presented in the 3rd row. For the Transducer baseline, the predictor layers are 2 instead of 4, since the model with 2 predictor layers gives the best results in our experiments.  
TAED outperforms the Transducer baseline except the dev-clean dataset, and J-TAED excels both Transducer and TAED in all 4 evaluation sets. Compared  J-TAED results with  TAED results, J-TAED achieves about 5.1 $\sim$ 10.5\% relative WER reduction (WERR)  with the help from extra text training data. 

\begin{table}
    \centering
    \caption{Recognition results for the J-TAED model on the \textsc{Librispeech} dataset.}
    \begin{tabular}{m{5em}|m{1cm}|m{1cm}|m{1cm}|m{1cm}}
    \toprule
      \multirow{2}{*}{model} & \multicolumn{2}{c|}{dev} & \multicolumn{2}{c}{test} \\
    \cline{2-5}
      &  clean & other & clean & other  %
    \begin{filecontents*}{tables/librispeech_main.csv}
model,devclean,devother,testclean,testother
Transducer, 2.09,5.50,2.26,5.50
TAED, 2.18,4.98,2.21,4.92
J-TAED, \textbf{1.95},\textbf{4.64},\textbf{2.09},\textbf{4.67}
%Mixformer Transducer, 2.09,5.50,2.26,5.50
%Conformer Transducer, 2.17,5.38,2.35,5.47
\end{filecontents*}
\csvreader[head to column names]{tables/librispeech_main.csv}{}%
    {\\\hline    \model &  \devclean & \devother & \testclean & \testother }
    \\
    \bottomrule
    \end{tabular}
    \label{tab:librispeech_rst}
\end{table}

As discussed in~\autoref{sec:multi-encoder}, we switch the Conformer layer to Transformer layer in the shared encoder to resolve the resolution discrepancy among speech and text modalities. 
\autoref{tab:multi_encoder} presents the Transducer results with Conformer encoder and multimodality encoder. 
Both encoders have 17 layers. The model size for the multimodality encoder is reduced from 120 million to 102 million since no convolution module is used in Transformer layers of the multimodality encoder. 
The average WERs for the Conformer and multimodality encoder based Transducer models are 3.83 and 3.84.  
The results are comparable though the multimodality encoder utilizes less parameters. 
We hypothesis that the top encoder layers are more focused on semantic information at utterance level and convolution modules play a less important role in those layers. It is aligned with the observation in natural language processing~\cite{Kudugunta2019InvestigatingMN,Cordonnier2020OnTR}. 

\begin{table}[t]
    \centering
    \caption{Comparison between Conformer encoder and Multimodality encoder on the \textsc{Librispeech} dataset.}
    \begin{tabular}{c|c|c|c|c|c}
    \toprule
    \multirow{2}{*}{model} & \#params & \multicolumn{2}{c|}{dev} & \multicolumn{2}{c}{test} \\
    \cline{3-6}
      & (m) & clean & other & clean & other  %
      \\
      \hline
      Conformer    &  120 & 2.17 & 5.38 & 2.30 & 5.47\\ % 3.83
      \hline
      Multimodality   & 102 & 2.09 & 5.50 & 2.26& 5.50 \\ % 3.838
      \bottomrule
    \end{tabular}
    \label{tab:multi_encoder}
\end{table}

\subsection{Text-based Domain Adaptation}

% \noindent\textbf{\textsc{SPGISpeech} dataset}~\autoref{tab:SPGISpeech_adaptation} presents results from J-TAED models before and after adapted to the out-of-domain \textsc{SPGISpeech} task. The transcription text of the \textsc{SPGISpeech} training data is used in the adaptation training. 
%  We evaluate models on two \textsc{Librispeech} dev sets and one \textsc{SPGISpeech} test set. The first two sets would be useful to monitor the degradation of the adapted model on the original domain that we collect our joint training data. One TAED and two J-TAED models are evaluated. 

\noindent\textbf{\textsc{SPGIspeech} dataset}~\autoref{tab:SPGISpeech_adaptation} presents the performances of TAED and two J-TAED models before and after adaptation in the out-of-domain \textsc{SPGISpeech} dataset. In all out-of-domain adaptation experiments, only the transcription text of the \textsc{SPGIspeech} training data is used instead of speech data. We evaluated models in two \textsc{Librispeech} development sets (clean and other) and one \textsc{SPGISpeech} test set. The first two sets would be useful to monitor the degradation of the adapted model in the original domain. 

The model in the 1st row in~\autoref{tab:SPGISpeech_adaptation} shows the results from the TAED model. The model in row 2 is the J-TAED model trained with \textsc{Librispeech} speech and the corresponding text training corpus. The model in row 3 (+Adapt) is adapted from the model in row 2 using \textsc{SPGIspeech} text data. This setup simulates a scenario where out-of-domain audio data is unavailable but text data exist only. The J-TAED model in row 4 (J-TAED $\bigstar$) is built with \textsc{Librispeech} speech, the corresponding \textsc{Librispeech} LM text data as well as \textsc{SPGIspeech} training text data. The number of transcription text from \textsc{SPGISpeech} is about half of the number of sentences from the \textsc{Librispeech} LM training data. In this scenario, out-of-domain text is available at the early model development stage (e.g. scratch model training). In addition, row 5 shows the results by applying text only adaptation of \textsc{SPGISpeech} text data to the row 4 model. This experiment evaluates whether further fine-tuning on same out-of-domain text still brings additional benefits or how it affects on the in-domain and out-of-domain performances.

% The experiments will help to answer following questions: 1) can we leverage out-of-domain text data in the early model training stage and 2) is additional text based adaptation helpful for such model? 

% The second and third rows are the results from the first J-TAED model before and after adaptation. 
% The J-TAED model is built with  \textsc{Librispeech} speech training corpus and corresponding language model training corpus. 
% %as shown in the second row of~\autoref{tab:SPGISpeech_adaptation}. The third row gives the corresponding results after text based adaptation. 
% The second J-TAED model includes the same training data as the first 
% J-TAED as well as transcription text from the \textsc{SPGISpeech} training set.  It tries to illustrate cases that we might only have target domain text but not speech data the  when we build the model. It will
% help to answer following two questions: 1) can we leverage those data in the early model training stage and 2) is additional text based adaptation helpful for such model? 
% The corresponding results before and after adaptation are presented in the last two rows. 

There are three observations from \autoref{tab:SPGISpeech_adaptation}. First, J-TAED achieves better results than TAED not only in the in-domain tasks but also the out-of-domain task as results shown in the row 1 and 2. 
Joint training might help to improve the model performance even at the out-of-domain task.
Second, as you can compare the 2nd and 3rd rows, adapting model with \textsc{SPGISpeech} text substantially lowers the WER of the \textsc{SPGISpeech} test set. Compared with the vanilla J-TAED model trained with \textsc{Librispeech} data in row 2, the adapted model has only minimal impact on the in-domain task (\textsc{Librispeech} ), where the WER is increased about 5\% WER. It indicates  text-based adaptation could effectively reduce WERs on the out-of-domain \textsc{SPGISpeech} task. This highlights the effectiveness of conducting text domain adaptation on J-TAED  and it can be a practical option especially when target or out-of-domain speech data is limited or unavailable. Third, integrating the target domain text in the joint training also helps to reduce the domain mismatch (row 4) and WER is reduced by about 5\% compared with the vanilla J-TAED in row 2.  
Additional text-based adaptation on the same text corpus could further alleviate domain mismatch but 
there is no guarantee the adaptation results from the model jointly trained with target 
domain text will be better (row 3 and row 5).

\begin{table}
    \centering
    \caption{WER for the text domain adaptation on the \textsc{SPGISpeech} dataset. $\bigstar$ indicates extra \textsc{SPGISpeech} text data is included in the model joint trained stage.} 
    % dev_clean | dev_other | SPGIspeech | inhouse |
    %                         ORIGN|
    \begin{tabular}{m{0.5cm}|m{5em}| m{1cm} | m{1cm} | m{1cm}}
    \toprule
     \multirow{2}{*}{id} &\multirow{2}{*}{model} & \multicolumn{2}{c|}{\textsc{Librispeech} dev} & {\textsc{SPGI}} \\
    \cline{3-4}
     && clean & other & \textsc{Speech} \\
     \hline
    1 &TAED & 2.18 & 4.98 & 14.31 \\ 
     \hline
    2 & J-TAED & 1.95 & 4.64 & 14.01 \\
    3 & + Adapt & 2.09& 4.94& 11.86 \\
     \hline
    4 & J-TAED $\bigstar$ & 1.95 & 4.64 & 13.27 \\
    5 & + Adapt & 2.12 & 4.79 & 12.35 \\
    \bottomrule
    \end{tabular}
    \label{tab:SPGISpeech_adaptation}
\end{table}
% (14.01 - 11.86)/14.01 * 100 = 15.3 
% (13.27 - 12.35)/13.27 * 100 = 6.9

\begin{figure}
    \centering
\begin{tikzpicture}
\pgfplotsset{compat=1.5}
\begin{axis}[
    width  = 0.52*\textwidth,
    height = 6cm,
    bar width=13pt,
    ybar,
    enlargelimits=0.15,
    legend style={at={(0.5,-0.15)},
      anchor=north,legend columns=-1},
    ylabel={WER},
    nodes near coords style={font=\scriptsize},
    symbolic x coords={dev-clean, dev-other,dev-ih, test-ih},
    xtick=data,
    nodes near coords,
    nodes near coords align={vertical},
    ]
\addplot coordinates {(dev-clean,2.18) (dev-other,4.98) (dev-ih,35.34)  (test-ih,34.75) };
\addplot coordinates {(dev-clean,1.95) (dev-other,4.64) (dev-ih,35.20)  (test-ih,34.22) };
\addplot coordinates {(dev-clean,2.27) (dev-other,5.1) (dev-ih,29.07)  (test-ih,27.98)};
%\addplot coordinates { (dev-other,4.6) (in-house,34.2)};
%\addplot coordinates {(dev-other,4.8) (in-house,31.3)};
\legend{TAED,J-TAED, J-TAED-adapt}
\end{axis}
\end{tikzpicture}
    \caption{Text-based domain adaptation on the In-House dataset. ``*-ih'' represents In-House data.}
    \label{fig:inhouse_adaptation}
\end{figure}
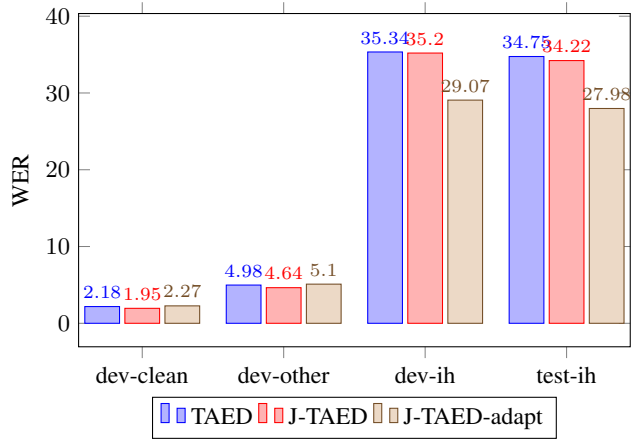

\noindent\textbf{In-House dataset}~\autoref{fig:inhouse_adaptation} demonstrates the J-TAED model adapted to an In-House dataset that is dominated with English named entities. Similar to the \textsc{SPGISpeech}
experiments, J-TAED outperforms than TAED in the In-House dataset though it is small. After adaptation, 
an average 17.8\% WER reduction is achieved in two In-House datasets at the cost about 10 $\sim$ 16\% WER increase in the original \textsc{Librispeech} dev sets.
%(35.2 - 29.07)/35.2 * 100 = 17.4 ;  ( 34.22 - 27.98)/34.22 * 100 = 18.2  ; (35.20 + 34.22 - 29.07 - 27.98 ) / (35.20 + 34.22) *100 = 17.8
 %(2.27-1.95)/1.95 = 16.4 , (5.1 - 4.64) / 4.64 = 9.9
%After text based adaptation, the WERs are reduced by 8.47\% and 8.39\% for two in-domain evaluation sets  respectively. 
 
%\begin{table}
%    \centering
%    \caption{Text domain adaptation recognition results} 
%    \begin{tabular}{c| m{1cm}  m{1cm} | m{1cm} m{1cm}}%| m{.8cm}}
%    \toprule
%     {Model} & {J-TAED} & {+text adaptation.} & {WERR}  \\
%     \hline
%     dev-clean &  1.95 &  2.01 &  2.04\%  %2.04
%    \\\cline{2-4}
%    dev-other &  4.64 &  4.83 &  4.61\%  % 4.61
%    \\\cline{2-4}
%    inhouse-val &  35.20 &  32.22 &  8.47\%  % (35.2 - 32.22)/35.2 * 100
%    \\\cline{2-4}
%    inhouse-test &  34.22 &  31.35 &  8.39\% % (34.22 - 31.35 ) / 34.22 * 100
%    \\
%    %  {Model} & {ModelA} & {+text adpt.} & {ModelB} & {+text adpt.} & {WERR}\\
%    %  \hline
%    %  dev-clean &  1.95 &  - &  - &  - & -\%  
%    % \\\cline{2-5}
%    % dev-other &  4.64 &  - & - & - & -\% 
%    % \\\cline{2-5}
%    % inhouse-val &  34.53 &  - & - & - & -\% 
%    % \\\cline{2-5}
%    % inhouse-test &  33.85 &  27.8 & - & - & -\% 
%    % \\
%    \bottomrule
%    \end{tabular}
%    \label{tab:inhouse_adaptation}
%\end{table}

\section{Conclusions}~\label{sec:conclusion}
In this work, a joint speech text optimization approach is proposed for TAED and is applied to ASR. Compared with the TAED baseline,  J-TAED achieves moderate improvement on the Librispeech data set with 5.8 $\sim$ 12.8\% WERR relative.
 %A fusion decoding method is proposed to leverage the advantages from both Transducer and AED decoding methods. 
 J-TAED  has unified multimodality representation for speech and text, and it is further extended to conduct text-based domain adaptation. 
 The adaptation results show that J-TAED can leverage the 
 linguistic information from the target domain text adaptation data and reduce the WER by 15.3\% and 17.8\% in the out-of-domain datasets. 

%\ifinterspeechfinal
%     The Interspeech 2025 organisers
%\else
%     The authors
%\fi
%would like to thank ISCA and the organising committees of past Interspeech conferences for their help and for kindly providing the previous version of this template.

\bibliographystyle{IEEEtran}
\bibliography{mybib}

\end{document}